# A new nature inspired modularity function adapted for unsupervised learning involving spatially embedded networks: A comparative analysis

Raj Kishore[1], Zohar Nussinov[2], Kisor K. Sahu[1]

1. School of Minerals, Metallurgical and Materials Engineering, Indian Institute of Technology Bhubaneswar- 752050, India
2. Department of Physics, Washington University in Saint Louis, MO- 63130-4899, USA

*Abstract*

Unsupervised machine learning methods can be of great help in many traditional science/ engineering disciplines, where huge amount of labeled data is not readily available or is extremely difficult/costly to generate. Two specific examples include the structure of granular materials and atomic structure of metallic glasses. While the former is critically important for several hundreds of billion dollars global industries, the latter is still a big puzzle in fundamental science. One thing is common in both the examples: the particles/atoms are the elements of the ensembles that are embedded in Euclidian space and one can create a spatially embedded network to represent their key features. Some recent studies show that clustering, which generically refers to unsupervised learning, holds great promise in partitioning these networks. In many complex networks, the spatial information (or, location) of nodes play very important role in determining the network properties. So understanding the structure of such networks is very crucial. We have compared the performance of our newly developed modularity function with some of the well-known modularity functions. We performed this comparison by finding the "best" partition in 2D and 3D granular assemblies. We show that for the class of networks considered in this article, our method produce much better results than the competing methods.

## 1. Introduction

Supervised machine learning is getting lots of attention these days primarily because of the success in deep learning, natural language processing, autonomous driving and robotics etc. [1-4]. If machine learning has to succeed in other traditional branches of engineering, probably lot of emphasis need to be put in unsupervised learning method also. This is so because, in those areas there are real scarcity in labeled data and without it, supervised learning is not going to have any meaningful impact. To illustrate this point in depth, here we will discuss two specific examples from material science. The first example involves granular materials. One of the central theme in material science is identifying the structure property correlation. While studying properties of granular materials at times might be tedious but generally not a big issue; the situation is complete opposite for describing its structure for which, as is argued in the following section, there is no satisfactory descriptor. This is despite its huge importance in diverse engineering disciplines such as: preparation of tablets by compacting powder ingredients in pharmaceutical industries; or understating the stress distribution in the silo walls when they need to be stored in a silo or transported through a conveyer; flow of chemicals through a packed bed reactor etc. [5-7]. Therefore, several hundred of dollars of global industries critically depend on it, but still there is no satisfactory descriptor for the structure of granular materials. Also, this is arguably the oldest science/engineering problem that human have ever attempted to study dating back to prehistorical era, as professor J. D. Bernal noted in his famous 1962 Bakerian lecture [8]: *"We are told that science is measurement but the first measure was the measure of heap –the measure of a heap of corn –by filling it into basket. Thus the measurement of volume preceded in this way the measurement both of length and weight. But in all the millennia that have passed in between, the real study of the nature of heaps has had to wait. It seemed too simple, in one respect, and too complicated in another."* In fact even today, still there is no satisfactory structural descriptor for granular materials. A scientifically logical way to study a complex problem is to deconstruct the problem first, and then study each individual pieces. Analogously, in this particular example, the first step will be to obtain a maximally optimal partition of the structure and then in the second step, to use a descriptor for each partition. Clustering, or the unsupervised learning method, hold a huge promise in addressing the first step [9-10] and this articles attempts to evaluate some of the suitable methods, though the list will not be exhaustive. Another example involves an area of very high fundamental importance, which is glass transition. When a molten metal alloy cools below some particular temperature also called as 'the glass transition point' at a high cooling rate (more than some 'critical cooling rate'), then the structure freezes to a glassy state, which can be practically termed as a 'solid' but technically correct term will be 'liquid' [11-12]. During this transition, the structure changes marginally (typically only by a few percentage), but properties like viscosity changes by a factor of $\sim 10^{13}$ or so, which cannot be explained by any accepted scientific theories. Although, evidence of some kind of structure-property relationship are provided by the existence of pentagonal symmetries [13], it is beyond doubt that a comprehensive picture is still amiss, primarily because of non-availability of any comprehensive descriptor for local structures. Now we contrast this situation with the case of standard metal alloys. Those standard alloys possess underlying atomic periodicity, which we call as crystal symmetries. The crystal symmetry act as an local as well as a global structural descriptor, which allows one to identify the location and amount of deviations from this idealization and use this knowledge effectively to study the property of interest for that particular material. Since, the glass is devoid of any such crystalline structural descriptor, the entire mechanism falls apart. The atomic ensemble can be considered as collection of nodal points embedded in a three-dimensional space. One can formulate the edges of the graph by using either the contacts between the neighboring atoms or by considering the force-transmitting atomic chains. Either way, one will be able to construct a network that is embedded in Euclidean space. Few previous works [14-17] have demonstrated that clustering, can be of great help to partition these spatial networks. In this particular study, we demonstrate considerable refinement over the existing protocols and compare them with some other methods proposed for unsupervised learning involving spatial networks. In the present article, we have compared the performance of different metrics through partitioning of a two dimensional (2D) and a three dimensional (3D) granular network.

## 2. Methods

### 2.1 Development of granular network

In this work, granular networks are created using two different protocols. The first protocol uses a physics-based Discrete Element Method (DEM) that simulates true granular dynamics (the detailed discussion can be found in [18]. It generates a 2D granular assembly of 7428 same size particles (Fig. 1(a)). The network can be constructed from this assembly by considering particle centers as node and an edge is considered between two particles if they are physically touching each other. The network have mostly 6-regular nodes (all nodes have degree six) separated by irregular (node degree is equal to or less than six) nodes (Fig. 1(b)). In the second protocol, we synthetically generate granular ensemble where, particles are positioned in a 3-D cubic crystal structure, and 50% particles from the central XY and XZ planes are removed to produce a patterned ordered region (Fig. 1(c)). This structure has four sparsely interconnected cuboidal boxes. The network generation from this assembly is same as in 2D assembly.

### 2.2 Different metrics used for clustering

Since there exist no universally accepted definition of "community", loosely one can define a community (or a 'module') as a group of nodes which have more interactions between them as compared to rest of the nodes [19-20]. Modularity optimization is the one of the first and simplest methods to detect these group of nodes. It maximizes the function known as modularity defined in [21].

$$Q = \frac{1}{2m} \sum_{ij} (A_{ij} - P_{ij}) \delta(\sigma_i, \sigma_j) \qquad (1)$$

Here summation goes over all possible node pairs $ij$ and $A_{ij}$ is the connection matrix ($A_{ij}=1$ if node $i$ and $j$ are connected else $A_{ij}=0$), $m$ is the total number of nodes in the network, $P_{ij}$ is the "Null" model used by different models. The value of kronecker delta function $\delta$ depends on the community membership of $i^{th}$ and $j^{th}$ node ($\sigma_i$ and $\sigma_j$). It gives zero value for different community membership and one for same membership

#### 2.2.1 Newman-Girvan (NG) modularity function

The NG modularity contrasts the number of actual connections in a given network to that of a null model. The NG modularity function uses configuration null model, which is obtained by randomly changing the network connections without altering degree of respective nodes. It compares the actual connection $A_{ij}$ to that with its probability in configuration null model $P_{ij}$. For single layer network, the NG modularity is given as

$$P_{ij} = \frac{k_i k_j}{2m} \qquad (2)$$

Where, $k_i$ and $k_j$ are the degree of $i^{th}$ and $j^{th}$ node. So from eq. (1) and (2), the modularity function based on configuration null model (NG function) is given as

$$Q_{NG} = \frac{1}{2m} \sum_{i,j} \left( A_{ij} - \frac{k_i k_j}{2m} \right) \delta(\sigma_i, \sigma_j) \tag{3}$$

### 2.2.2 Gravitational Null model

In the previous section, Newman-Girvan uses configuration null model $P_{ij} = \frac{k_i k_j}{2m}$ which do not incorporated the spatial information of nodes. It gives same value for a node pair with degree $k_i$ and $k_j$ irrespective of whether they are near or far from each other. Whereas in many real life network, proximity plays very important role in edge probability between nodes. So the inclusion of proximity effect in community detection in such networks becomes crucial. Expert et al. [22] proposed a spatial null model which assumes that the interaction between two nodes is directly proportional to their importance and decays with distance. The gravity null model is given as

$$p_{ij}^{gr} = I_i I_j f(d_{ij}) \tag{4}$$

Where, $f(d)$ is the 'deterrence function' used to defines the effect of space on node interactions. $I_i$ and $I_j$ are the node importance. There are different ways to calculate the node importance. The simplest choice can be the strength/degree of the node $I_i = K_i = \sum_j w_{ij}$. It will make this null model similar to NG null model. The deterrence function for weighted networks is formulated as

$$f(d_{ij}) = \frac{\sum_{\{k,l | d_{kl} = d_{ij}\}} W_{kl}}{\sum_{\{k,l | d_{kl} = d_{ij}\}} (I_k I_l)} \tag{5}$$

Above equation is applicable for discrete data sets but distance between node pairs is a continuous data, so binning of distance data set is required. It ensures that there will be enough nodes for each distance bin thus one can construct a meaningful deterrence function $f(d)$ in eq. (5). There may be different binning method possible. We have used equal distance binning method. So from eq. (4) and (5) gravitational null model can be written as

$$p_{ij}^{gr} = I_i I_j \frac{\sum_{\{k,l | d_{kl} = d_{ij}\}} W_{kl}}{\sum_{\{k,l | d_{kl} = d_{ij}\}} (I_k I_l)} \tag{6}$$

Thus the gravitational null model based modularity function can be written as

$$Q_{Gr} = \frac{1}{2m} \sum_{i,j} \left( A_{ij} - p_{ij}^{gr} \right) \delta(\sigma_i, \sigma_j) \tag{7}$$

### 2.2.3 Radiation Null model

Gravitational null model included the spatial information of the nodes by adding a deterrence function. This function is not always non-zero as NG function. In many spatially embedded networks, there might be different fluxes between locations that are at same distance. For example the spread rate of an infectious disease from one location to other will be highly affected by the population density between these two locations. The gravitational null model is unable to capture these differences in fluxes.

The radiation model [23] is developed to address these issues. It was mainly focused on population flow and successfully applied in several situations [24-25]. The radiation null model used mean commuting flux between locations $i$ and $j$ with populations $n_i$ and $n_j$ given as

$$T_{ij} = T_i \frac{n_i n_j}{(n_i + r_{ij})(n_i + n_j + r_{ij})} \tag{8}$$

Where, $T_i$ is the number of commuters at location $i$ and $r_{ij}$ is the population between location $i$ and $j$. The radiation null model is given as

$$p_{ij}^{rad} = \hat{T}_{ij} \frac{\sum_{\{k,l|d_{kl}=d_{ij}\}} W_{kl}}{\sum_{\{k,l|d_{kl}=d_{ij}\}} \hat{T}_{kl}} \tag{9}$$

$$\hat{T}_{ij} = \frac{(T_{ij} + T_{ji})}{2} \tag{10}$$

So from eq. (1) and (9), the modularity function based on radiation null model can be formulated as

$$Q_{rad} = \frac{1}{2m} \sum_{i,j} (A_{ij} - p_{ij}^{rad}) \delta(\sigma_i, \sigma_j) \tag{11}$$

### 2.2.4 New modularity function

This method is based on optimization of modularity function (Eq. 12). This function is developed for spatially embedded network and can be utilized for several applications [26-28]. The detailed description of this function can be found in [29].

$$Q(\sigma) = \frac{1}{2m} \sum_{i,j} (a_{ij} A_{ij} - \theta(\Delta x_{ij}) |b_{ij}| J_{ij})(2\delta(\sigma_i, \sigma_j) - 1) \tag{12}$$

Here m, is the total edges, $\theta(\Delta x_{ij})$ is Heaviside unit step function used to restrict the neighborhood definition by defining $\Delta x_{ij} = x_c - |\vec{r}_i - \vec{r}_j|$ as the difference between cutoff distance $x_c$ and Euclidian distance between nodes $i$, $j$. The $a_{ij}$ and $b_{ij}$ are the strength (not to be confused with edge weights) of connected and missing edges between $i^{th}$ and $j^{th}$ nodes respectively. The strength $a_{ij}$ and $b_{ij}$ can be calculated in different ways based on the type of network. Here since we are dealing with unweighted networks, the strength $a_{ij}$ and $b_{ij}$ can be calculated as

$$a_{ij} = b_{ij} = \left( \frac{\bar{k}_i + \bar{k}_j}{2} - <k> \right) \tag{13}$$

where, $\bar{k}_i = \frac{1}{n_i + 1} \left( \sum_{j=1}^{n_i} k_j + k_i \right)$ is the average degree of neighborhood (number of nodes inside the cutoff distance) of $i^{th}$ node, $<k> = \frac{1}{N} \sum_{r=1}^{N} k_r$ is the average degree of the network and $k$ is the node degree and $N$ is total number of nodes. The cutoff distance $x_c$ is chosen based on the fact that in 2D and 3D there can never be an edge between particle with its second nearest neighbor. As the closest packed structure in 3D is face centered cubic (FCC) and in 2D, hexagonal packing, we have chosen the distance between second nearest neighbor in these structure as the cutoff distance $x_c$.

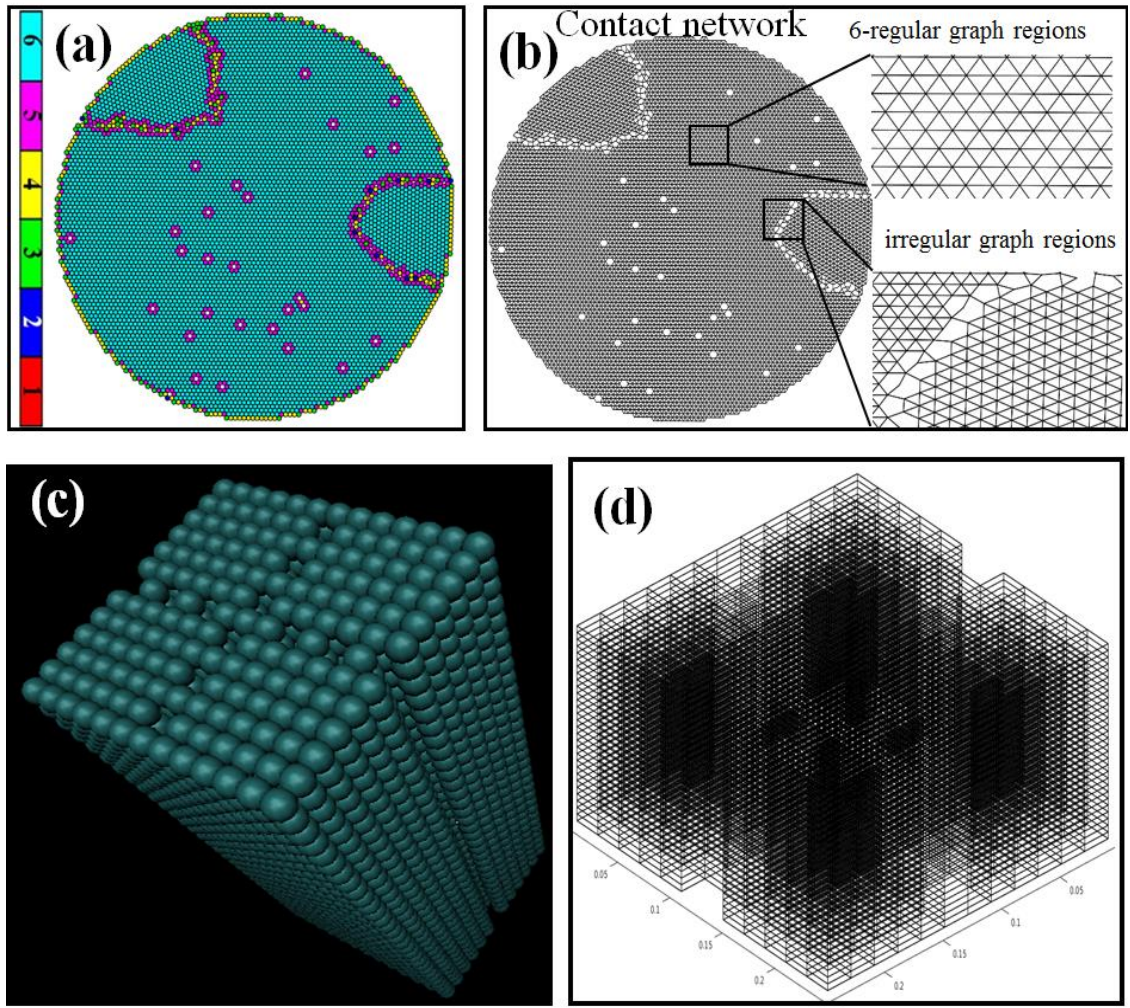

**Figure 1:** Granular ensemble and corresponding granular network (a) Packing of 7428 uniform sized disks obtained by centripetal packing through DEM simulation (ref). The particles are colored based on their coordination numbers (see the colorbar at the extreme left) and (b) its corresponding network (c) 3-D cubic crystal structure, where only 50% particles from the XY and XZ planes are removed and (h) its corresponding granular network

## 2.3 Clustering algorithm

The algorithm that we have used tries to merge a node with any of its connected neighbors in order to increases the overall modularity of the network. This process continues until no more merging is possible *i.e.* none of the merging increases the modularity of the network. This process rapidly converges towards the local maxima in the modularity curve. The pictorial representation of the algorithm is given in Fig 2.The algorithm consist of mainly three steps:

*(i)    Initialization:* In this step, the connection matrix which is also called as adjacency matrix is calculated and edge weights to all the edges are assigned. All the nodes are assigned as single node community. So initially the number of community is same as the number of nodes in the network. This type of initial community distribution is called symmetric distribution. Let us say this partition has modularity $Q_0$. There are other ways also to initialize the number of communities for e.g. we can constrained the number communities as *q*. But this might not give the natural division of the network and has not been adopted in this study.

*(ii)    Node merging:* In this step each node is selected iteratively. The selected node is then checked for possible merging sequentially with all its connected neighbour communities and modularity is calculated for each case (for e.g. *Q1, Q2, Q3* etc.: refer "node merging" in Fig. 2(c)). Finally node will be merged in the community in which the difference in the modularity of network after and before merging is highest positive. This steps continues until no further merging increases the modularity of the system.

*(iii)    Community merging:* The node merging method used in step (ii) often converged at local maxima in the modularity curve. In order to remove any local maxima trap in the modularity curve, we merge any two connected community. If this merging increases the modularity then we keep these merging else reject it.

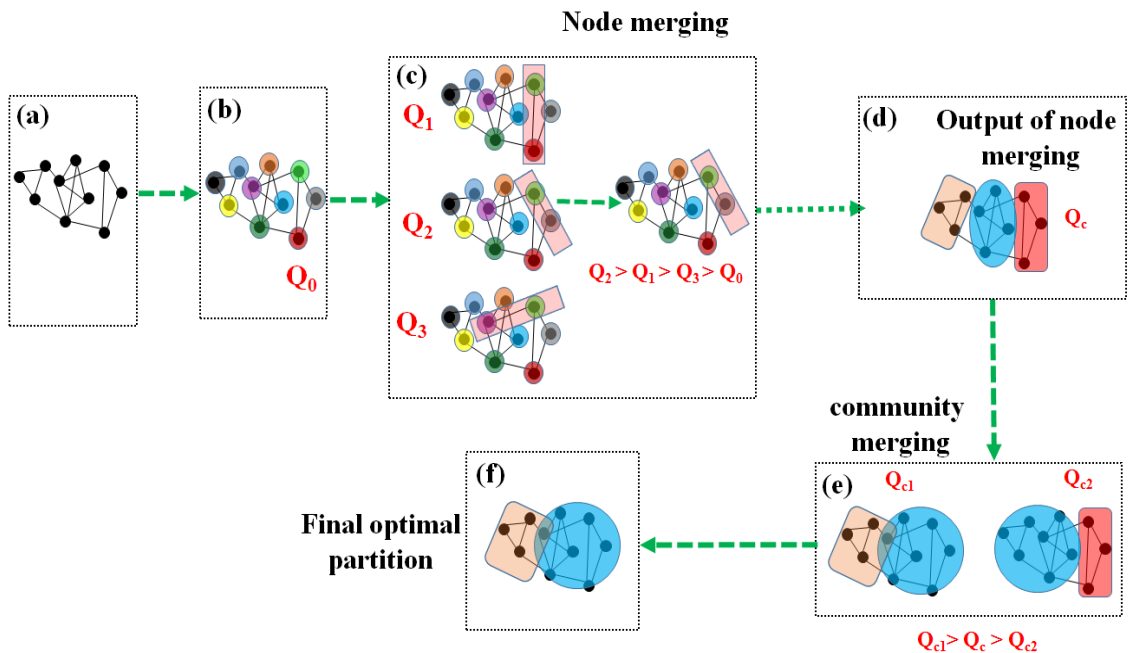

Figure 2: Pictorial representation of algorithm of network clustering using modularity maximization. (a) The given network. (b) Initially all the nodes are assigned different community membership (different color). (c) In node merging, each node community is iteratively merged with its connected neighbor node

community and merged with the node which increases maximum modularity of the system. (d) After all the node pairs are merged (no further merging increases the modularity), (e) community merging is performed and (f) finally optimal partition is obtained. Here Q refers to the modularity of the partitioned network.

## 3. Results

We have partitioned the granular assembly of 7428 uniform size particles (Fig. 1(a)) using different modularity functions discussed in "*Method*" section. The corresponding network obtained from this granular assembly (Fig. 1(b)) has mostly ordered regions where the degree of all nodes are same and equals to six (in two dimensional assembly of same size particles, maximum first touching neighbor can be six). These ordered regions are separated through disordered regions having degree less than or equal to six. The given granular assembly has three such ordered regions and, we considered this as the feature of this assembly. Since the ordered regions have no inhomogeneity, a good clustering algorithm should not partition these regions into more than one cluster *i.e.* the entire one ordered regions should belongs to one cluster. The partitioned granular assembly by all the four different modularity functions are shown in (Fig. 3). The proposed modularity function has clustered the three big ordered regions of the given network into separate and single communities (Fig. 3(b)). It has also clustered the inhomogeneity of the given network into separate communities. Whereas the Newman modularity functions has partitioned the three big ordered regions into many communities (Fig. 3(c)). This is a kind of false partitioning and unable to extract any useful structural information from the given network.

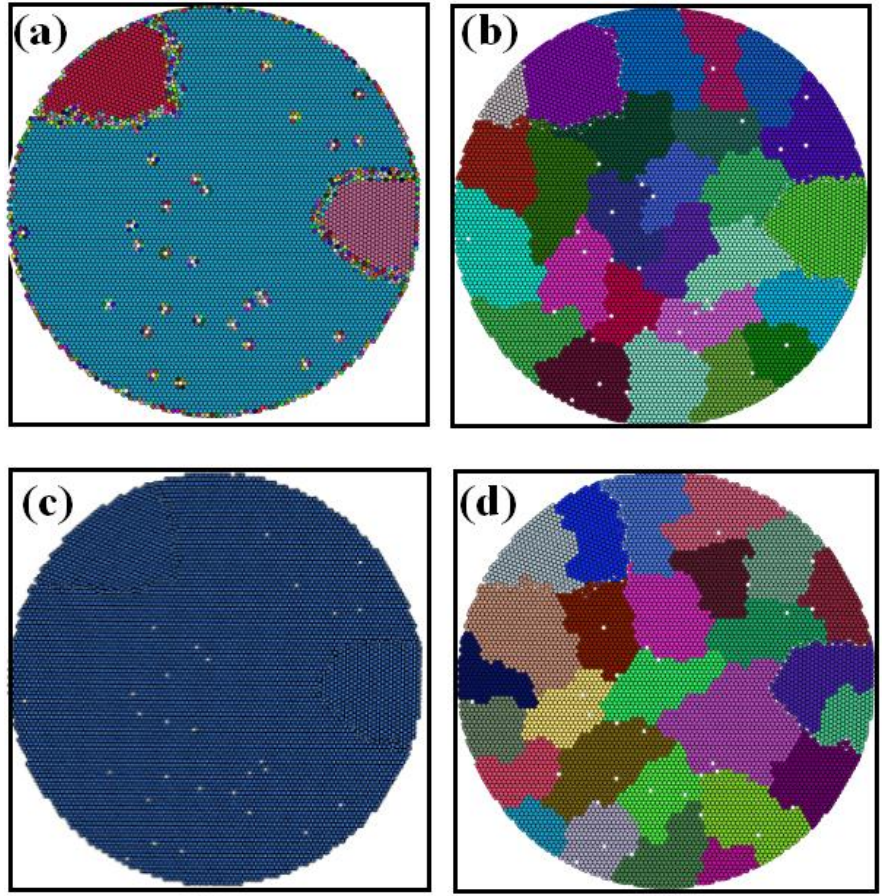

Figure 3: Partition of 2D granular assembly shown in Fig 1(a) by using different metrics. (a) Partition obtained by proposed function, (b) Newman function, (c) Gravitation function, (d) Radiation function

Similarly the gravitational function has clustered the entire network into one community (Fig. 3(d)). This is called trivial solution of clustering and conveys no information about the structural features of the network. The radiation function has over-partitioned the given network (Fig. 3(e)). It has partitioned the ordered regions into smaller communities, which is not desired. It is not clear how the function is deciding the cluster boundaries inside the ordered regions. So it is reasonable to argue that in clustering, the proposed modularity function is outperforming as compared to the other existing modularity functions.

We have also compared the performance of theses modularity functions by partitioning the network obtained from 3D assembly shown in Fig. 1(c). This assembly has four cuboidal blocks which contains similar type of particles (based on the coordination number).

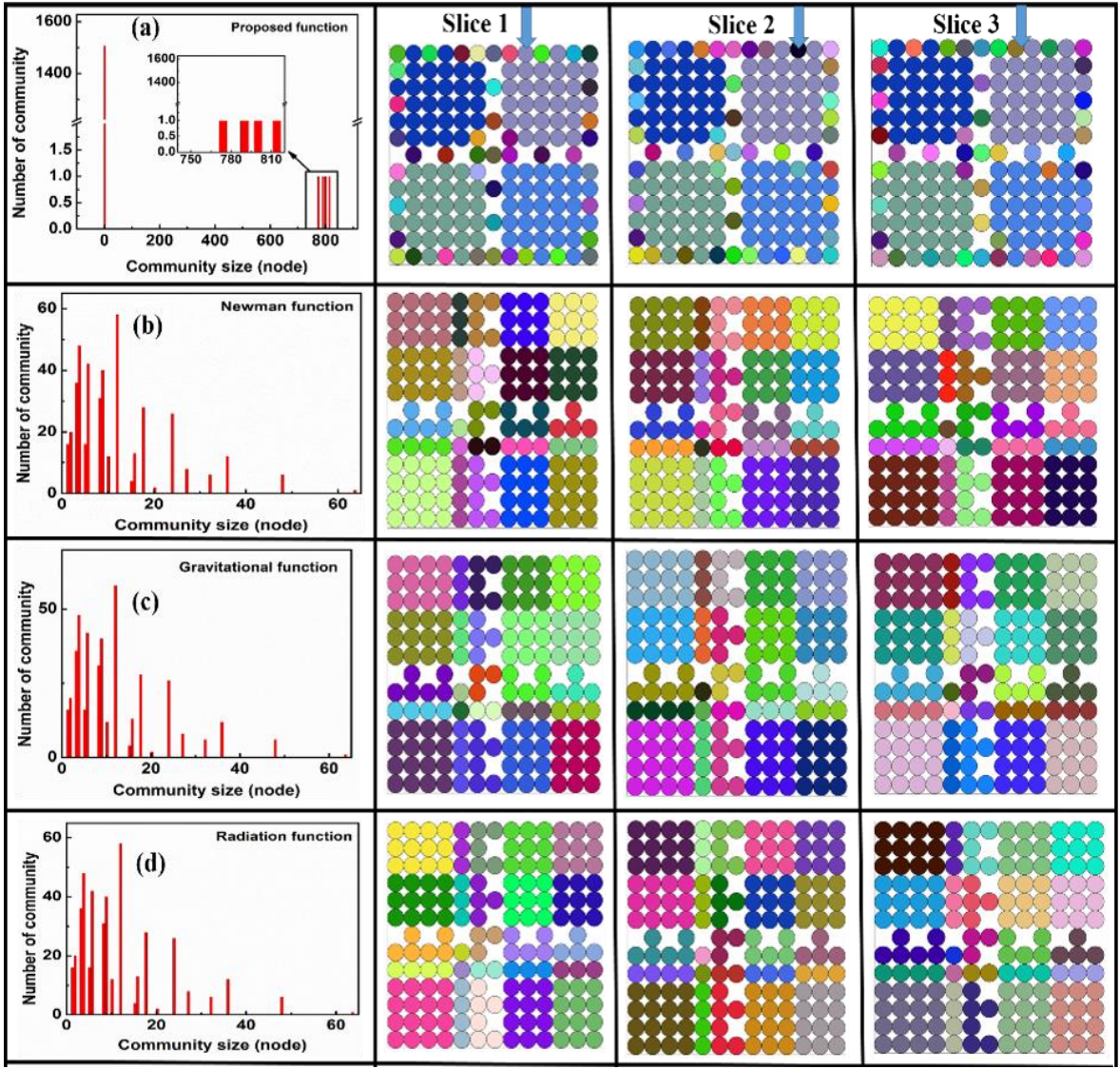

Figure 4: Partition of 3D granular assembly shown in Fig. 1(c) by using different metrics. (a) The community size distribution of the partitioned structure obtained by using proposed function (b) Newman function (c) Gravitation function (d) Radiation function. The corresponding community structure at three XY planes (slice 1-3) are shown in corresponding rows.

Ideally these four blocks should be identified as four separate communities. The community size distribution of these partitioned network is shown in Fig. 4 (first panel). It shows that except for the new modularity function (Fig. 4(a)), the other three modularity functions are over-partitioning these blocks. Whereas the new modularity function is clearly finding four communities as shown in inset figure of Fig. 4(a). In order to visualize these over-partitioned community structures, we need to look into the 3D structure. For this purpose, we have selected three random XY slices (2D) at different Z positions (shown in Fig. 1(c)). In order to show visually different communities, the nodes are colored based on their community membership. The community structure of these three slices are shown next to the community size distribution curve in the same panel (Fig. 4a - d). In the partition obtained by new modularity function (Fig. 4(a)), the color assigned to the four big communities (four cuboidal blocks) remains consistent for all the three slices. This shows that these four blocks are divided into four separate communities and not over-partitioned. Whereas for other three modularity functions (Fig. 4(b-d)), the color changes from one slice to other and even in same slice also. This indicates that the partition obtained through these three modularity function is over-partitioned.

**Conclusions**

In the present article, we have partitioned the given 2D and 3D network using different modularity functions. Our modularity function has detected correctly the three big ordered regions of the given 2D granular assembly. Whereas other metrics have either over-partitioned the ordered regions or merged them in one community. Not only for 2D assembly, but the new modularity function has produced much better partition in case of 3D assembly also as compared to other three modularity functions. In both the studied assemblies, our proposed modularity function is capable of detecting their structural features.

**Acknowledgement**

KKS was partially funded by UAY project (84/IITBBS-006) and ZN was partially funded by NSF grant number 1411229.